\documentclass[final,5p,times,twocolumn]{elsarticle}

\usepackage{amssymb}
\usepackage[font={scriptsize,it}]{caption}
\usepackage[font=small,labelfont=bf]{caption}
\usepackage[font=scriptsize]{subfig}
\usepackage{enumerate}
\usepackage{amsmath}
\usepackage{url}
\usepackage{lineno}
\usepackage{multirow}
\usepackage{color}

\usepackage{hyperref}

\journal{Elsevier}

\begin{document}

\begin{frontmatter}

\title{HEp-2 Cell Classification: The Role of Gaussian Scale Space
Theory as \\A Pre-processing Approach}

\author{Xianbiao Qi}
\ead{qixianbiao@gmail.com}
\author{Guoying Zhao}
\ead{gyzhao@ee.oulu.fi}
\author{Jie Chen}
\ead{jiechen@ee.oulu.fi}
\author{Matti Pietik{\"a}inen}
\ead{mkp@ee.oulu.fi}
\address{Center for Machine Vision Research, University of Oulu, PO Box 4500, FI-90014, Finland.}



\begin{abstract}
\textit{Indirect Immunofluorescence Imaging of Human Epithelial Type 2} (HEp-2) cells is an effective way to identify the presence of Anti-Nuclear Antibody (ANA). Most existing works on HEp-2 cell classification mainly focus on feature extraction, feature encoding and classifier design. Very few efforts have been devoted to study the importance of the pre-processing techniques. In this paper,
we analyze the importance of the pre-processing, and investigate the role of Gaussian Scale Space (GSS) theory as a pre-processing approach for the HEp-2 cell classification task. We validate the GSS pre-processing under the Local Binary Pattern (LBP) and the Bag-of-Words (BoW) frameworks.
Under the BoW framework, the introduced pre-processing approach, using only one Local Orientation Adaptive Descriptor (LOAD), achieved superior performance on the Executable Thematic on Pattern Recognition Techniques for Indirect Immunofluorescence (ET-PRT-IIF) image analysis. Our system, using only one feature, outperformed the winner of the ICPR 2014 contest that combined four types of features. Meanwhile, the proposed pre-processing method is not restricted to this work; it can be generalized to many existing works.
\end{abstract}




\end{frontmatter}



\section{Introduction}
\label{sec1}
\textit{Indirect Immunofluorescence Imaging of Human Epithelial Type 2} (HEp-2) cells~\cite{foggia2013benchmarking, foggia2014pattern} is a commonly used way to identify the presence of Anti-Nuclear Antibody (ANA) that is considered as an effective approach to diagnose various autoimmune diseases. Before, the human experts are required to identify the types of HEp-2 cells according to their experience. This process is highly subjective depending on the experience of the experts, and errors usually happen especially when considering the large intra-class variations and small inter-class variations in the HEp-2 cells.  The recognition of HEp-2 cells is a typical pattern recognition problem.
Recently, several contests held in the past three years on the HEp-2 cell classification have greatly raised the interests in the development of an effective recognition system. There were 28, 14, and 11 submissions individually submitted to the ICPR 2012~\cite{icpr2012contest}, ICIP 2013~\cite{icip2013contest, hobson2015} and ICPR 2014~\cite{icpr2014contest} HEp-2 cell classification contests. Those methods range from applying fast morphological methods~\cite{ponomarev2014ana}, to designing or transferring new features or feature encoding approaches or classifiers~\cite{ersoy2012hep}, and to fusing different approaches~\cite{theodorakopoulos2012hep, manivannan2014hep}.

Existing works on the HEp-2 cell classification mainly focused on three aspects: feature extraction, feature encoding and classifier. Among all three aspects, the feature extraction received most of the attention. Many well-known features were applied to this application, such as Scale Invariant Feature Transformation (SIFT)~\cite{lowe2004distinctive}, Local Binary Pattern (LBP)~\cite{ojala2002multiresolution} and Gray Level Co-occurrence Matrix (GLCM)~\cite{haralick1973textural}. Meanwhile, there were also some new features proposed for the task, such as Co-occurrence of Adjacent LBP (CoALBP)~\cite{nosaka2014hep} and Shape Index Histogram (SIH)~\cite{larsen2014hep}. The feature encoding is an important stage in the traditional Bag-of-Words (BoW)~\cite{csurka2004visual} model. Many advanced feature encoding approaches, such as Hard Assignment, Kernel Codebook, Sparse Coding (SC), Local-constrained Linear Coding (LLC) and Improved Fisher Vector (IFV), were studied on this task. A novel feature encoding technology named as Fisher Tensor~\cite{faraki2014fisher} was also proposed. The choices of classifiers are important to the final classification accuracy. The Support Vector Machine (SVM)~\cite{cortes1995support} is the most widely used classifier on this task. There were also some works studying the effectiveness of other classifiers, such as Shareboost~\cite{ersoy2012hep}, K-NN~\cite{stoklasa2014efficient}.

By contrast, very few efforts have been devoted to study the importance of the pre-processing technique.  We highly agree that the above mentioned three aspects are important on the HEp-2 cell classification, but we also believe that effective pre-processing technique will benefit this task greatly. Thus, in this paper, we propose to analyze the importance of the pre-processing, and investigate the role of Gaussian Scale Space (GSS) theory as a pre-processing technique. We propose to evaluate the GSS pre-processing in two different frameworks: the LBP framework and the BoW framework. Extensive experiments show that the proposed GSS pre-processing technique greatly boosts the recognition performance of the approach without using the GSS in both frameworks.

One submission based on the proposed method achieved superior performance (with Mean-Class-Accuracy (MCA) 84.63\%) on the ET-PRT-IIF\footnote{\url{http://mivia.unisa.it/iif2014/}}. Using only one type of feature, our approach greatly outperformed the winner of the ICIP 2013 contest that combined two features, and exceeded the winner of ICPR 2014 contest that combined four types of features. The source code submitted to the contest is provided at the link 
\footnote{\url{https://www.dropbox.com/s/q7xuht2ddwgr81f/PRLettersMaterial.zip?dl=0}}.




\subsection{Related Works}
\label{sec2}

Existing works on the HEp-2 cell classification can be categorized into three categories:

{\bf{Feature Extraction. }}In pattern recognition tasks, the feature extraction is always one of the most important stages. It greatly affects the final classification performance. On the HEp-2 cell classification, the LBP~\cite{ojala2002multiresolution} and many of its variants have been applied to this task, such as Completed LBP (CLBP)~\cite{guo2010completed}, Co-occurrence of Adjacent LBP (CoALBP)~\cite{nosaka2014hep}, Pairwise Rotation Invariant Co-occurrence of LBP (PRICoLBP)~\cite{qi2014pairwise}. In the ICPR 2012 contest, the CoALBP ranked 1st among all 28 submissions. Later, in the ICIP 2013 contest, a combination of the PRICoLBP and the bag of SIFT won the contest among all the 14 submissions. In the recently held ICPR 2014 contest, an ensemble of four features~\cite{manivannan2014hep} (Multi-resolution Local Patterns, RootSIFT, Random Projections and Intensity Histograms) achieved the best Mean-Class-Accuracy (MCA) among all the 11 submissions on the Task 1 (cell classification). The same method also achieved the 1st rank on the Task 2 (specimen classification)~\cite{manivannan2014hepSpecimen}.
According to the previous three contests, it is easily found that, until now, the feature extraction dominates the task.

{\bf{Feature Encoding. }}In the BoW framework, the feature encoding is an important stage. Since the BoW model was widely used on the HEp-2 cell classification, there were several works~\cite{kong2014hep, xu2014linear, ensafi2014bag} that focus on transferring advanced feature encodings (the reader can refer to the evaluation presented in~\cite{chatfield2011devil} to get a detailed information about some advanced encoding methods.) or designing new encoding techniques. For instance, in the ICIP 2013 contest, Shen et al.~\cite{icip2013contest} used hard assignment for the Bag of SIFT model. In the ICPR 2014 contest, Ensafi et al. \cite{ensafi2014bag} used Sparse Coding (SC) to encode the SIFT and SURF features, and Manivannan et al.~\cite{manivannan2014hep} used the Local-constrained Linear Coding (LLC)~\cite{wang2010locality} to encode four types of features.  Recently, Faraki et al.~\cite{faraki2014fisher} introduced a novel encoding approach called Fisher Tensors for the HEp-2 cell and texture classification.

{\bf{Classifier. }}The classification stage is the last stage of the whole system, thus it is directly related to the recognition performance. Nearest Neighbor (NN) classifier is the simplest classification method. It does not require any training. But the evaluation will become very slow when the scale of the problem is large. In ~\cite{stoklasa2014efficient}, Stoklasa et al. proposed to mine efficient K-NN for the task. Ersoy et al.~\cite{ersoy2012hep} proposed to use the Shareboost to conduct the classification. Most submissions in the previous contests used the linear or kernel SVM. Usually, the SVM shows better performance than the NN classifier. We believe some other classifiers are also worth studying in future, such as Random Forests, Gaussian Processes.


\subsection{Contributions}
The novelty of this paper focuses on analyzing the role of the pre-processing. We propose to use a multi-resolution Gaussian Scale Smoothing to pre-process the image before the feature extraction stage. For the GSS,

\begin{itemize}
\item we visually explain the underlying reasons why the GSS pre-processing can greatly benefit the HEp-2 cell classification task.
\item we experimentally show the significant improvement brought by the proposed pre-processing approach. One submission based on the proposed method to the ET-PRT-IIF achieves superior performance on this task.
\end{itemize}

The remainder of this paper is organized as follows. In Section \ref{sec:mainbody}, we introduce the proposed GSS pre-processing approach, and analyze the underlying reasons that make the GSS work on the HEp-2 cell classification. In Section \ref{sec:experiments}, we evaluate the proposed approach in detail and compare it the state-of-the-art approaches. In Section \ref{sec:conclusion}, we conclude this paper with a discussion.



\section{The Role of Gaussian Scale Space Theory}
\label{sec:mainbody}

\begin{figure}[h]
\includegraphics[width=0.49\textwidth]{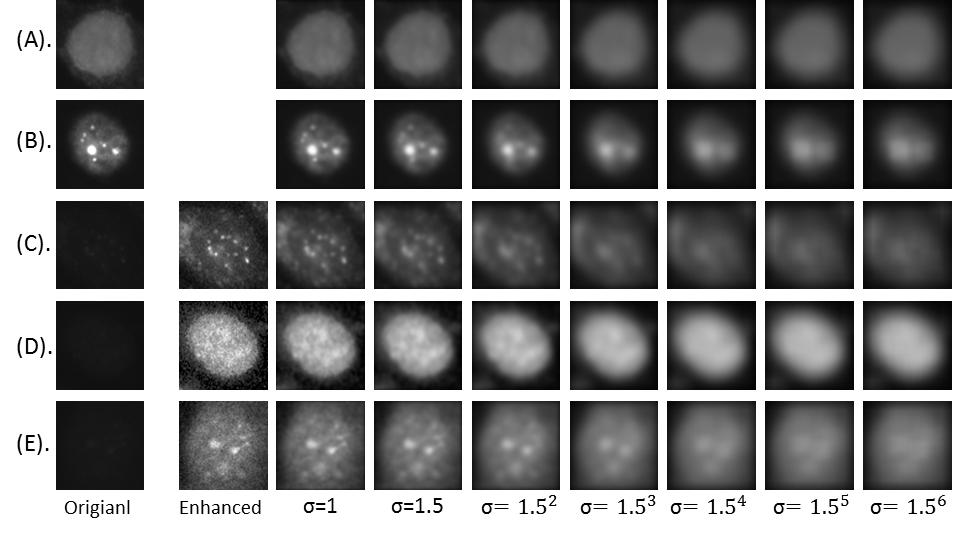}
\centering
\caption{Filtered images under different Gaussian scales. (A) and (B) belong to the ``Positive'' type, and (C), (D) and (E) come from the ``Intermediate'' type of HEp-2 cells. The ``Intermediate'' cell images are enhanced for better visualization.}
\label{fig:gssscalesFig}
\end{figure}

\subsection{Gaussian Scale Space Theory}
Scale Space Theory (SST) \cite{lindeberg1993scale} is a multi-scale signal representation framework. Given
a 2-D image $I(x, y)$, the scale space of the image can be defined as:
\begin{equation}\nonumber
\label{eq:gss}
    L(x, y, {\sigma}) = F(x, y, {\sigma}) * I(x, y),
\end{equation}
where $*$ is the convolution operation in $x$ and $y$, $F(x, y, \sigma)$ is a filtering function, and $\sigma$ is the scale factor.

In the SST, Gaussian function is the most widely used filtering function. It was shown by \cite{koenderink1992surface, lindeberg1993scale} that the Gaussian function is the only possible scale-space kernel under a variety of reasonable assumptions. Thus, in this paper, we will use the Gaussian function as the scale-space kernel. This is termed as Gaussian Scale Space (GSS) in literature. The 2-D Gaussian filter can be defined as:
\begin{equation}\nonumber
\label{eq:gssfilter}
    G(x, y, {\sigma}) = \frac{1}{2\pi {\sigma}^2} \exp(-\frac{x^2+y^2}{2{\sigma}^2}).
\end{equation}


In this paper, we propose to regard the GSS as a pre-processing technique for the HEp-2 cell classification task. Before, the feature extraction stage, we use the GSS to pre-smooth each image and then extract features from the filtered images.


{\bf{Analysis of underlying reasons}} why the GSS as a pre-processing works for the HEp-2 cell classification task. From our viewpoint, we believe the following four aspects may explain the reasons:
\begin{itemize}
\item Remove noise. As shown in Figure~\ref{fig:gssscalesFig}, the HEp-2 cells in the data set have strong noise, especially on the ``Intermediate'' images. Since the ``Intermediate'' images are hardly visible to human eyes in their original status, thus, we use a simple image enhancement~\footnote{$I(x, y) = \frac{I(x, y) - minI}{maxI-minI}$, where $minI$ and $maxI$ are the minimum and maximum values of $I$ individually.} algorithm to enhance the images. We can see from the enhanced ``Intermediate'' images that there is strong noise in the ``Intermediate'' cells. In this situation, the GSS as a smoothing pre-processing as shown in Figure 1 can effectively remove the noise.
\item Enhance texture information. With the increase of the Gaussian scale, the filtered image will become more smooth. In this way, the global texture information will tend to dominate the structure. With multi-resolution filtering, the subsequent features can capture cross-scale texture and shape information. Thus, the discriminative power of the features will be enhanced.

\item Boost the discriminative power of the final image representation by increasing the number of features. Suppose we pre-process the image with $K$ Gaussian scales, then we will get $K$ filtered images. Densely sampling features from the filtered images, the number of features increases by a factor of $K+1$ times compared to the direct sampling in the original images. In this way, multi-scale features are sampled, and the discriminative power of the final image representation are enhanced.

\item Ease the misalignment of the scales among different images under the Bag-of-Words framework. Due to some potential reasons, such as the camera is partially out of focus or the slight density variation of serum dilution~\footnote{The ICPR 2014 contest data set used a serum dilution of 1:80.}, the image scale may vary. Bagging of features extracted from the multi-resolution GSS filtering is an effective way to ease this problem.
\end{itemize}


\subsection{Gaussian Scale Space Theory as a Pre-processing}

In this paper, we study the GSS Theory as a pre-processing in two different frameworks: the LBP framework~\cite{ojala2002multiresolution} and the BoW model~\cite{csurka2004visual}.

\begin{figure}[h]
\includegraphics[width=0.50\textwidth]{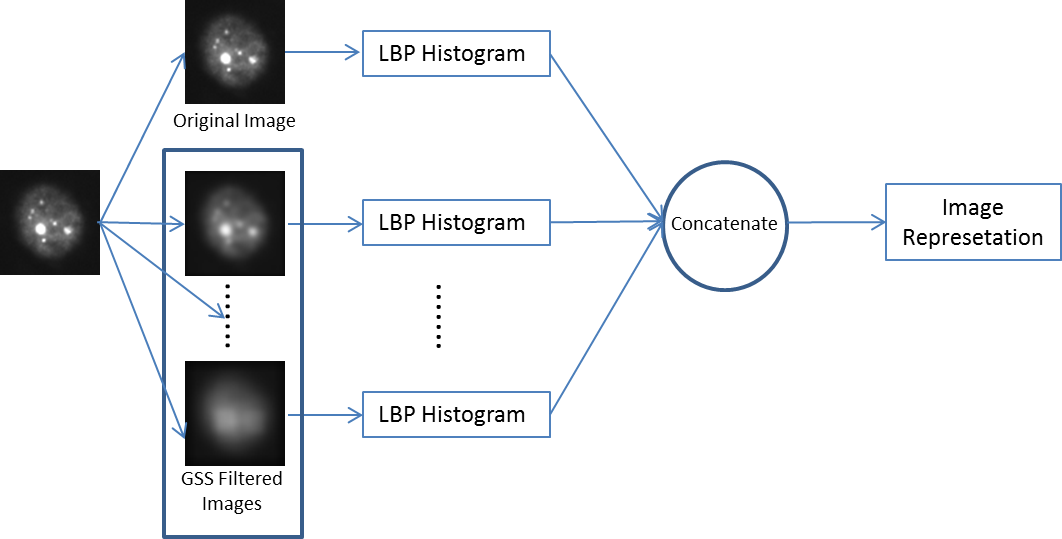}
\centering
\caption{The flow chart of image representation under the LBP framework with the Gaussian Scale Space theory as a pre-processing.}
\label{fig:lbpframework}
\end{figure}

{\bf{GSS as a Pre-processing in the LBP model.}} The flow chart of Scale Space Theory as a pre-processing in the LBP framework is shown in Figure~\ref{fig:lbpframework}. As shown in Figure~\ref{fig:lbpframework}, the input image is filtered with Gaussian filters in different scales. Then, the LBP histogram is extracted from each filtered image. Finally, all LBP histograms are concatenated into the final image representation.

To build a LBP histogram for each image, the LBP pattern in each pixel should be firstly computed as:
\begin{equation}
LBP({n, r}) = \sum_{k=0}^{n-1} s(g_k-g_c)2^k, \ \ \ \ \
s(x) = \begin{cases}
1, & x \ge 0 \\
0, & x < 0, \\
\end{cases}
\end{equation}
where $n$ is the number of neighbors and $r$ is the radius, and $s(.)$ is sign function. $g_c$ is the gray value of the central pixel, and $g_k$ is the pixel value of its $k$-th neighbor. After obtaining LBP pattern for each pixel, a histogram can be built. Multi-scale strategy can be used to enhance the discriminative power of the descriptor by choosing different $(n, r)$. In practice, to control the dimension of final representation, a rotation invariant uniform LBP (RIU-LBP) is used. The dimension of the RIU-LBP equals to $n+2$.
Suppose the feature dimension extracted from one image is $D$, the final dimension of image representation that concatenates the features from the original image and $K$ filtered images is $(K+1)\times D$.

Most LBP variants can follow this framework including the traditional LBP, the Completed LBP (CLBP)~\cite{guo2010completed}, the Co-occurrence of Adjacent LBP (CoALBP)~\cite{nosaka2014hep} and Pairwise Rotation Invariant Co-occurrence of LBP (PRICoLBP)~\cite{qi2014pairwise}. In this paper, instead of pursuing higher performance, our aim is to demonstrate the effectiveness of our proposed pre-processing method, thus, we do not use these advanced LBP variants.

\begin{figure*}[t]
\includegraphics[width=1.0\textwidth]{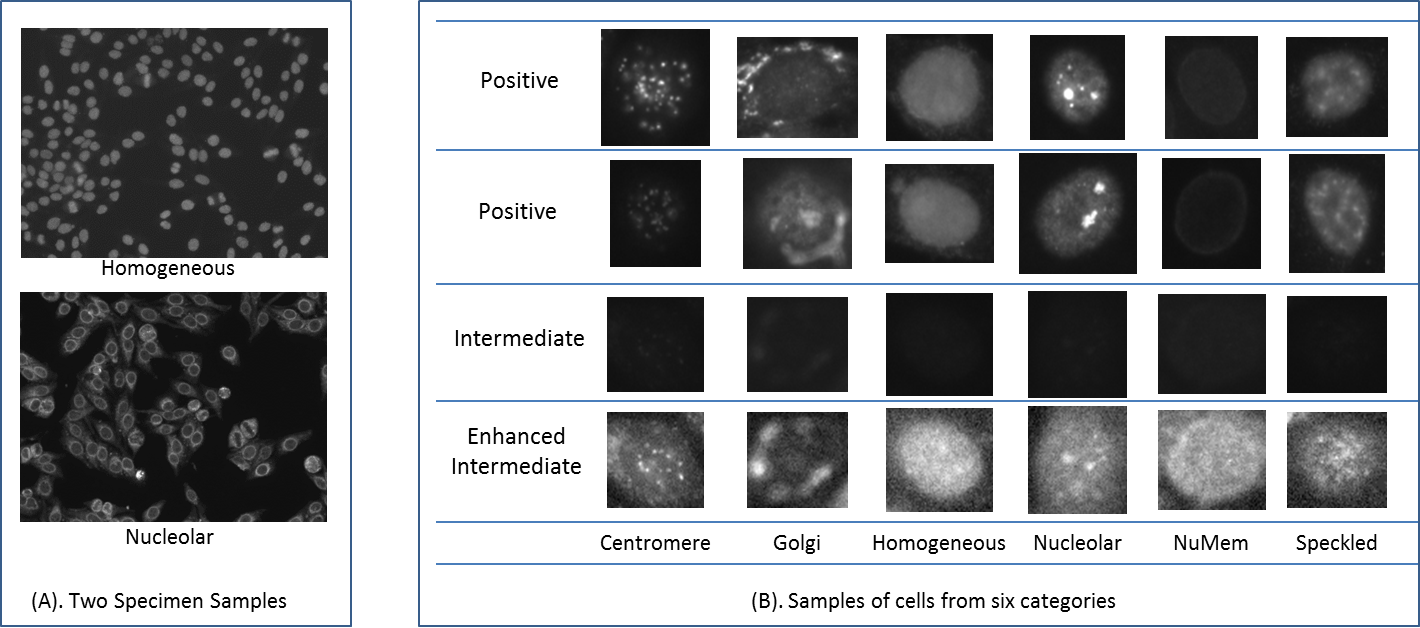}
\centering

\caption{Examples of the specimen and cells. On the left panel of the figure, we show two specimen images. On the right panel, three samples for all six categories are shown, in which the first two rows show the ``Positive'' type and the third row shows one ``Intermediate'' type. The fourth row shows the corresponding enhanced images for the third row. Easy to see that the intra-class variation is big especially when considering the ``Positive'' and ``Intermediate'' types in the same category.}
\label{fig:samples33}
\end{figure*}

\begin{figure}[h]
\includegraphics[width=0.46\textwidth]{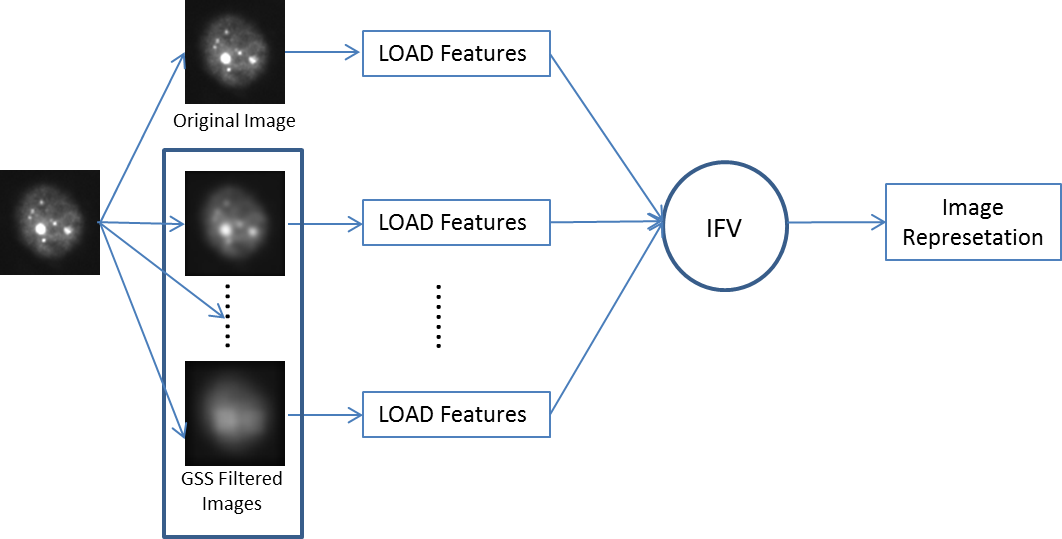}
\centering
\caption{The flow chart of image representation under the BoW framework with the Gaussian Scale Space theory as a pre-processing.}
\label{fig:bowframework}
\end{figure}


{\bf{GSS as a Pre-processing in the BoW model.}}
The flow chart of the BoW model using the GSS pre-processing is shown in Figure~\ref{fig:bowframework}. First, each input image is filtered with Gaussian filters of different scales. Then, local features, such as the Local Orientation Adaptive Descriptor (LOAD) \cite{qi2015nc},  can be densely extracted from the original image and all the filtered images. Finally, all LOAD features extracted from all scales are pooled into one IFV encoding to form the final image representation. 
The strategy of pooling all feature of different scales into one IFV representation can well ease the misalignment of the scales among different images. The BoW framework with the GSS pre-processing is different from the LBP framework with the GSS pre-processing. In the LBP framework, we extract one LBP histogram from each filtered image and concatenate all LBP histograms into the final representation, but in the BoW framework, only one BoW histogram is constructed for the features extracted from all filtered images. 

The choice of local features is flexible. All popular features, such as SIFT~\cite{lowe2004distinctive}, SURF~\cite{bay2006surf}, MORGH~\cite{fan2012rotationally} and LIOP~\cite{wang2011local} can be used. In the paper, we used our newly proposed LOAD. In nature, the LOAD descriptor can be considered as the LBP built on an adaptive coordinate system. In LOAD, we used four scales (LBP(8, 1), LBP(8, 2), LBP(8, 3), LBP(8, 4)). In each scale, only the uniform LBP patterns (59 patterns) were used. Therefore, the feature dimension of the LOAD was $59\times 4 = 236$.
With filtered images in different scales, the features extracted from patches can capture multi-resolution information.
As pointed out before, the features extracted from the filtered images with large scales can capture more global texture information, while in the original or filtered images with small scales, the feature can depict fine detailed information.

The proposed pre-processing approach also does not have any requirement to the feature encoding. Any feature encoding can be applied in the subsequent processing, such as the Vector Quantization (VQ), Soft Assignment, Kernel Codebook, Sparse Coding (SC), Local-constrained Linear Coding (LLC) and Improved Fisher Vector (IFV). In this paper, we use the IFV due to its effective encoding ability. 

The IFV representation measures the average first and second order differences between the local features and the Gaussian Mixture Models (GMMs). First, the Principal Component Analysis (PCA), such as $D$ components, is used to remove the correlation between arbitrary two dimensions. Then, the GMMs, such as $K$ GMMs, are learned from the after-PCA features. The average first and second order differences of the after-PCA features w.r.t. to the GMMs parameters are calculated and concatenated. Therefore, the final dimension of the IFV representation is $2\times D \times K$.
The IFV proves to be effective when only using linear SVM. The linear SVM can greatly facilitate the final evaluation stage. The computational cost of the IFV is also low. We refer the reader to read the references~\cite{chatfield2011devil, perronnin2010improving} to get a more through understanding of the IFV.

\begin{figure*}[t]
\includegraphics[width=1.0\textwidth]{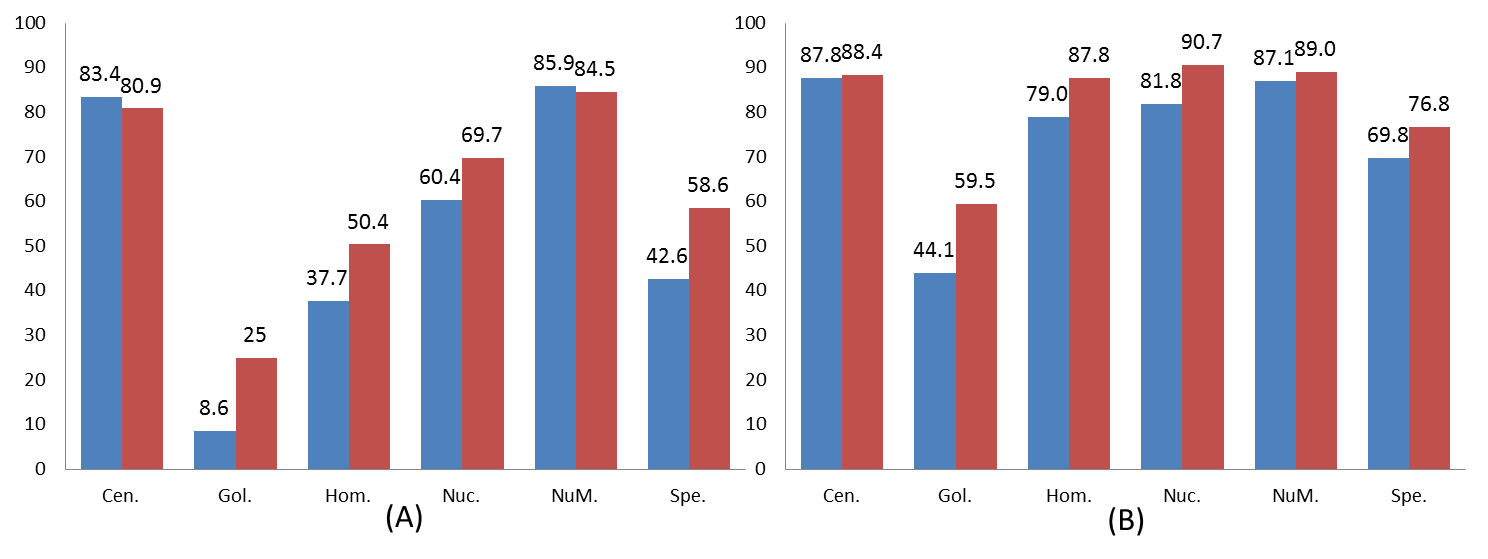}
\centering
\caption{Evaluation under (A). the LBP framework, and (B). the BoW framework. Methods without using GSS pre-processing are marked with blue color and methods with GSS preprocessing are marked with red color.}
\label{fig:LBPBoWFig}
\end{figure*}

\section{Experimental Analysis}
\label{sec:experiments}
\subsection{Dataset, Implementation Details and Evaluation Strategy}
{\bf{Dataset.}} The I3A-2014 Task-1 data set was collected between 2011 and 2013 at the Sullivan Nicolaides Pathology laboratory, Australia. The whole data set consists of 68,429 cells coming from six categories. The six classes are: Homogeneous (2,494 cells from 16 specimens), Speckled (2,831 cells from 16 specimens), Nucleolar (2,598 cells from 16 specimens), Centromere (2,741 cells from 16 specimens), Golgi (724 cells from 4 specimens), and Nuclear Membrane (2,208 cells from 15 specimens). The I3A-2014 Task-1 of the ICPR 2014 contest used the same data set as the previous ICIP 2013 contest.

The training part contains 13,596 cell images that are collected from 83 different specimen images.
The testing part consists of 54,833 cell images. The test data is privately maintained by the organizers and not publicly available until now. All results evaluated on the test data set were reported by the contest organizers. Two specimen images from the I3A-2014 Task-2 are shown in the left panel of Figure \ref{fig:samples33}, and some cells from each class are shown in the right panel of Figure \ref{fig:samples33}.


\begin{figure*}[t]
\includegraphics[width=1.0\textwidth]{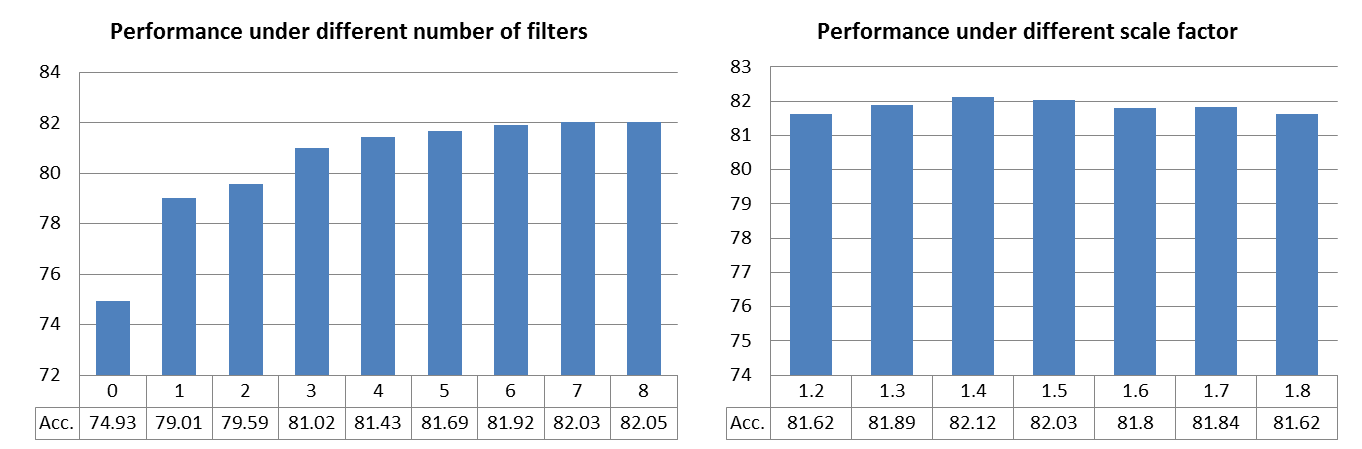}
\centering
\caption{Evaluations of the parameters. Left panel: classification accuracy under different number of filters. Right panel: classification accuracy under different scale factors.}
\label{fig:Parameters}
\end{figure*}


{\bf{Implementation Details.}} We evaluate the GSS as a pre-processing step in two different ways: the LBP and the BoW frameworks. For both methods, we use the original image and seven filtered images ($\sigma = {b}^{n-1}, b=1.5$ and $n=1,2...7$) in default. We will evaluate the influence of different scale factor $\sigma$ and different number of filters below.

In the LBP approach, we use three scales ((8, 1), (16， 2) and (24, 3)), and use rotation invariant uniform LBP. Therefore, the dimension of the feature vector extracted from one image is 54. Since we concatenate all features from original image and the filtered images, thus the dimension of the final feature vector is $(1+7)\times 54= 432$. This framework is extremely fast, it takes less than 0.2s to process one image on a desktop with dual-core 3.4G CPU.

In the BoW framework with Improved Fisher Vector (IFV) encoding, we densely extract the LOAD features\footnote{Detailed information about the LOAD can be found at \cite{qi2015nc}.} from circular patches with the radius 13 with a stride of two pixels in $y$-axis and one pixel in $x$-axis.
On an image of size $70\times 70$, around 4,600 LOAD features will be extracted.
For the IFV, we use Principal Component Analysis (PCA) to decrease the dimension to 100 and then use 256 Gaussian Mixture Models (GMMs) to cluster the after-PCA features. Thus, the dimension when using one dictionary is $2\times 100 \times 256 = 51,200$. Detailed description of the IFV can be found in ~\cite{perronnin2010improving, chatfield2011devil}. For our submission, to improve the stability of our algorithm, we use two dictionaries. But in our experiments, we only observed slight improvement (0.07 percentage point) with two dictionaries compared to using only one dictionary under the Leave-One-Specimen-Out strategy.
The whole system takes less than 1.6s to process an image (including around 1.0s for feature extraction of the LOAD, 0.6s for feature encoding, and almost none of time for classification because of using linear SVM.).
For the linear SVM, we use the Liblinear~\cite{fan2008liblinear} to train and evaluate the model. For the IFV, we use the Vlfeat~\cite{vedaldi2010vlfeat} toolbox.

Two {\bf{Evaluation Strategy}} are used in the paper:
\begin{itemize}
\item Leave-One-Specimen-Out (LOSO) Evaluation. In the LOSO strategy, cell images from any 82 specimens are used for training, and the rest cell images from one specimen is used for testing. The final Mean-Class-Average (MCA) is reported based on the 83 splits. The strategy is an effective way to evaluate the algorithm when the test data set is not available.
\item Evaluation on the test data set. Evaluation on the test data set is a fair way to evaluate different algorithms. Every submission is blind to the test data. Meanwhile, the scale of the test data is large.
\end{itemize}


\subsection{Comparative Analysis of Gaussian Scale Space}
To evaluate the effectiveness of the GSS as a pre-processing, we conduct two sets of experiments, one uses the GSS pre-processing and the other one does not. We use the LOSO evaluation strategy.
The category-wise accuracies using the LBP framework or the BoW framework are shown in Figure~\ref{fig:LBPBoWFig}(A), and Figure~\ref{fig:LBPBoWFig}(B).

We can find from the Figure~\ref{fig:LBPBoWFig} that:
\begin{itemize}
  \item In both frameworks, the GSS pre-processing significantly improves the performance of that without pre-processing. In the LBP framework, the GSS boosts the MCA by around 8 percentage points. In the BoW framework, the GSS improves the MCA by around 7 percentage points.
  \item On some categories, such as ``Golgi'', ``Homogeneous'' and ``Speckled'', the GSS pre-processing under both frameworks greatly boosts the performance.
\end{itemize}


\subsection{Evaluation of Different Number of Filters and Different Scale Factor}
In this subsection, we evaluate the influence of the parameters to the classification performance under the BoW framework. The evaluation is conducted to answer two issues: 1). How many scales should be used? 2). What is the optimal scale factor $\sigma$? For the first question, we evaluate the BoW model under 9 different configurations; the results are shown in the left panel of the Figure~\ref{fig:Parameters}. For instance, ``0'' means we only use the features extracted from the original image, and ``n'' means we use the features extracted from the original image and $n$ filter images with filter factors from ${1.5}^{1-1}$ to ${1.5}^{n-1}$. For the second question, we evaluate the BoW model under 7 different $b$ ($\sigma = b^{n-1}, n=1,2,...,7$) configurations; the results are shown in the right panel of the Figure~\ref{fig:Parameters}.

From the left panel of the Figure~\ref{fig:Parameters}, we have two findings. First, the performance of using the pre-processing significantly improves that without using the pre-processing. For instance, only using one filtered image improves the performance by 4.1 percentage points rather than that of without using filtered image. Second, the performance almost saturates when around seven filters are used; using more filters does not bring in performance gain, but increases the computational cost. Therefore, in the following experiments, we used seven filtered images.

From the right panel, the proposed approach works best when $b$ is set around 1.4; the differences between different configurations are small. The following results were based on the setting $b=1.5$ because we used this setting in our previous submission to the contest.

%
%
%

\subsection{Experimental Results using the LOSO Strategy on the Training Data Set}
Since the test data set is not provided, we found that the leave-one-specimen-out strategy is an effective way to evaluate different methods. The same strategy was also used in some previous works (Vestergaard et al. ~\cite{larsen2014hep} and Manivannan et al. ~\cite{manivannan2014hep}). In this strategy, the cells from 82 specimens among 83 specimens are used for training and the rest cells from one specimen are used for testing. The results are based on average of 83 splits.
The category-wise accuracy of three different approaches and classification confusion matrix of our approach are shown in Table ~\ref{tab:confusingmatricsLeaveoneout}. The results in the Table ~\ref{tab:confusingmatricsLeaveoneout} reveal that:

\begin{table}[!h]
  \centering
  \caption{The category-wise accuracy of different approaches and classification confusion matrix of our BoW with the LOAD feature using the leave-one-specimen-out strategy on the training data.}
  \subfloat[Category-wise classification accuracy.]{
    \small
    \centering
    \begin{tabular}{ @{\ }c@{\ } | @{\ }c@{\ } | @{\ }c@{\ } | @{\ }c@{\ } | @{\ }c@{\ } | @{\ }c@{\ } | @{\ }c@{\ } | @{\ }c@{\ } }
    \hline
        \% & Cen. & Gol. & Hom. & Nuc. & NuM. & Spe. & Average\\
        \hline
        IFV  & 87.81& 44.06 & 79.03  & 81.83 & 87.09 & 69.83 &74.93\\[0.03cm]
        Vestergaard & 85.04& 50.97 & 84.88 & 87.49 & 88.81 & 75.03 &78.70\\[0.03cm]
        Manivannan  & 85.66& 58.01 & 81.8  & 90.65 & 88.04 & \bf{77.36} &80.25\\[0.03cm]
        \bf{GSS\_IFV (Ours)}          & \bf{88.43} &\bf{59.53} & \bf{87.77} & \bf{90.69} & \bf{88.99} & 76.76 & \bf{82.03}\\[0.03cm]
        \hline
        \end{tabular}
  }

  \subfloat[Classification confusion matrix of our approach(\bf{82.03}).\label{tab:chapter4:1c}]{
    \small
    \centering

    \begin{tabular}{|@{\ }l@{\ } | @{\ }l@{\ } | @{\ }l@{\ } | @{\ }l@{\ } | @{\ }l@{\ } | @{\ }l@{\ } | @{\ }l@{\ }|}
    \hline
        \% & Cen. & Gol. & Hom. & Nuc. & NuM. & Spe. \\
        \hline \hline
        Cen. & \bf{88.43} & 0.26        & 0.73       & 1.610     & 0          & 8.97\\[0.03cm]
        Gol. & 3.73       & \bf{59.53}  & 6.63       & 19.89     & 7.87       & 2.35\\[0.03cm]
        Hom. & 0.08       & 0.28        & \bf{87.77} & 1.04      & 1.56       & 9.26\\[0.03cm]
        Nuc. & 2.39       & 0.96        & 1.62       & \bf{90.69}& 0.85       & 3.50\\[0.03cm]
        NuM. & 0          & 1.77        & 4.62       & 2.04      & \bf{88.99} & 2.58\\[0.03cm]
        Spe. & 8.55       & 0.07        & 8.27       & 4.95      & 1.41       & \bf{76.76}\\[0.03cm]
        \hline
        \end{tabular}
  }
\label{tab:confusingmatricsLeaveoneout}
\end{table}

\begin{itemize}
  \item Among all three algorithms, our method performs best, the approach of Manivannan et al. ranks 2nd and the method of Vestergaard et al. ranks 3rd. Note that Vestergaard et al.'s and our approach only use one type of feature, while Manivannan et al. combine four types of features.
  \item In all three methods, our approach achieves the highest performance on five categories. For instance, our approach improves the Vestergaard et al. by around 3 percentage points and outperforms the method of Manivannan et al. by around 6 percentage points on the category ``Homogeneous''. We believe the huge improvement on the ``Homogeneous'' accounts for (1): The proposed GSS pre-processing is effective and (2): The LOAD feature is good at capturing the texture information that is important in the category ``Homogeneous''.
  \item The category ``Golgi'' is easy to be misclassified into ``Nucleolar'', and the categories ``Homogeneous'' and ``Speckled'' are usually misclassified into each other. This may be because that the confusing pairs have similar texture and shape structures.
\end{itemize}



\begin{table*}[t]
\caption{Classification confusion matrices of seven different approaches on the test data set reported by the contest organizer.}\label{tab:chapter4:1}
  \centering
  \subfloat[ICIP2013\_Shen(\bf{80.70})\label{tab:chapter5:1a}]{
    \scriptsize
    \scriptsize
    \scriptsize
    \centering
    \begin{tabular}{@{\ }l@{\ } | @{\ }l@{\ } | @{\ }l@{\ } | @{\ }l@{\ } | @{\ }l@{\ } | @{\ }l@{\ } | @{\ }l@{\ }}
        \% & Cen. & Gol. & Hom. & Nuc. & NuM. & Spe. \\[0.1cm]
        \hline \hline
        Cen. & \bf{95.13} & 0.38 & 0.40 & 1.63 & 0.46 & 2.00\\[0.1cm]
        Gol. & 0.89 & \bf{60.05} & 9.67 & 14.76 & 12.23 & 2.40\\[0.1cm]
        Hom. & 0.39 & 0.76 & \bf{78.15} & 4.92 & 6.64 & 9.14\\[0.1cm]
        Nuc. & 1.13 & 1.18 & 1.70 & \bf{90.31} & 2.33 & 3.36\\[0.1cm]
        NuM. & 0.19 & 0.94 & 4.63 & 1.47 & \bf{90.85} & 1.92\\[0.1cm]
        Spe. & 10.39 & 0.82 & 13.40 & 3.24 & 2.47 & \bf{69.68}\\[0.1cm]
        \end{tabular}
  }
  \subfloat[ICIP2013\_Vestergaard(\bf{81.22})\label{tab:chapter5:1b}]{
    \small
    \scriptsize
    \centering
    \begin{tabular}{@{\ }l@{\ } | @{\ }l@{\ } | @{\ }l@{\ } | @{\ }l@{\ } | @{\ }l@{\ } | @{\ }l@{\ } | @{\ }l@{\ }}
        \% & Cen. & Gol. & Hom. & Nuc. & NuM. & Spe. \\[0.1cm]
        \hline \hline
        Cen. & \bf{96.21} & 0.30 & 0.25 & 1.34 & 0.46 & 1.44\\[0.1cm]
        Gol. & 0.39 & \bf{62.25} & 4.47 & 8.39 & 22.53 & 1.97\\[0.1cm]
        Hom. & 0.30 & 0.63 & {77.34} & 6.11 & 8.46 & 7.15\\[0.1cm]
        Nuc. & 1.17 & 1.00 & 1.68 & \bf{92.76} & 1.09 & 2.31\\[0.1cm]
        NuM. & 0.22 & 0.78 & 2.96 & 1.93 & \bf{92.10} & 2.01\\[0.1cm]
        Spe. & 11.28 & 0.65 & 14.85 & 4.33 & 2.22 & \bf{66.67}\\[0.1cm]
        \end{tabular}
  }
  \subfloat[ICPR2014\_Paisitkriangkrai(\bf{81.55})\label{tab:chapter5:1d}]{
    \small
    \scriptsize
    \centering
    \begin{tabular}{@{\ }l@{\ } | @{\ }l@{\ } | @{\ }l@{\ } | @{\ }l@{\ } | @{\ }l@{\ } | @{\ }l@{\ } | @{\ }l@{\ }}
        \% & Cen. & Gol. & Hom. & Nuc. & NuM. & Spe. \\
        \hline \hline
        Cen. & \bf{94.96} & 0.12 & 0.18 & 1.63 & 0.61 & 2.51\\[0.1cm]
        Gol. & 0.92 & \bf{65.11} & 4.21 & 15.13 & 11.97 & 2.66\\[0.1cm]
        Hom. & 0.13 & 0.21 & \bf{76.83} & 6.61 & 7.56 & 8.67\\[0.1cm]
        Nuc. & 1.02 & 0.51 & 1.17 & \bf{92.92} & 2.56 & 1.82\\[0.1cm]
        NuM. & 0.15 & 0.65 & 4.89 & 1.42 & \bf{91.08} & 1.80\\[0.1cm]
        Spe. & 13.59& 0.23 & 11.91 & 3.25 & 2.60 & \bf{68.42}\\[0.1cm]
        \end{tabular}
  }
  \hfill
  \subfloat[ICPR2014\_Gao(\bf{83.23})\label{tab:chapter5:1d}]{
    \small
    \scriptsize
    \centering
    \begin{tabular}{@{\ }l@{\ } | @{\ }l@{\ } | @{\ }l@{\ } | @{\ }l@{\ } | @{\ }l@{\ } | @{\ }l@{\ } | @{\ }l@{\ }}
        \% & Cen. & Gol. & Hom. & Nuc. & NuM. & Spe. \\
        \hline \hline
        Cen. & \bf{96.03} & 0.18 & 0.05 & 1.50 & 0.48 & 1.76\\[0.1cm]
        Gol. & 0.03 & \bf{73.20} & 5.75 & 10.42 & 9.14 & 1.45\\[0.1cm]
        Hom. & 0.19 & 0.84 & \bf{78.29} & 5.97 & 7.52 & 7.20\\[0.1cm]
        Nuc. & 0.72 & 1.33 & 1.86 & \bf{93.72} & 1.17 & 1.22\\[0.1cm]
        NuM. & 0.08 & 0.83 & 4.22 & 0.73 & \bf{91.27} & 2.87\\[0.1cm]
        Spe. & 11.31 & 0.59 & 14.61 & 4.80 & 1.85 & \bf{66.85}\\[0.1cm]
        \end{tabular}
  }
  \subfloat[ICPR2014\_Theodorakopoulos(\bf{83.33})\label{tab:chapter5:1c}]{
    \small
    \scriptsize
    \centering
    \begin{tabular}{@{\ }l@{\ } | @{\ }l@{\ } | @{\ }l@{\ } | @{\ }l@{\ } | @{\ }l@{\ } | @{\ }l@{\ } | @{\ }l@{\ }}
        \% & Cen. & Gol. & Hom. & Nuc. & NuM. & Spe. \\
        \hline \hline
        Cen. & \bf{94.74} & 0.25 & 1.31 & 1.68 & 0.15 & 1.87\\[0.1cm]
        Gol. & 0.30 & \bf{71.03} & 5.03 & 7.53 & 15.65 & 0.46\\[0.1cm]
        Hom. & 0.00 & 0.98 & \bf{74.31} & 3.36 & 13.21 & 8.14\\[0.1cm]
        Nuc. & 0.84 & 0.92 & 1.60 & \bf{92.85} & 2.24 & 1.54\\[0.1cm]
        NuM. & 0.17 & 1.46 & 3.64 & 1.11 & \bf{91.99} & 1.63\\[0.1cm]
        Spe. & 8.18 & 0.59 & 12.16 & 1.69 & 2.30 & \bf{75.08}\\[0.1cm]
        \end{tabular}
  }
  \subfloat[ICPR2014\_Sansone(\bf{83.64})\label{tab:chapter5:1d}]{
    \small
    \scriptsize
    \centering
    \begin{tabular}{@{\ }l@{\ } | @{\ }l@{\ } | @{\ }l@{\ } | @{\ }l@{\ } | @{\ }l@{\ } | @{\ }l@{\ } | @{\ }l@{\ }}
        \% & Cen. & Gol. & Hom. & Nuc. & NuM. & Spe. \\
        \hline \hline
        Cen. & \bf{95.52} & 0.42 & 0.21 & 1.15 & 0.05 & 2.66\\[0.1cm]
        Gol. & 0.03 & \bf{71.82} & 4.74 & 7.27 & 14.60 & 1.55\\[0.1cm]
        Hom. & 0.05 & 0.80 & \bf{78.57} & 4.94 & 8.07 & 7.58\\[0.1cm]
        Nuc. & 0.75 & 1.58 & 1.96 & \bf{92.55} & 1.70 & 1.46\\[0.1cm]
        NuM. & 0.05 & 0.76 & 3.14 & 0.85 & \bf{93.39} & 1.81\\[0.1cm]
        Spe. & 13.35 & 0.71 & 11.11 & 2.65 & 2.17 & \bf{70.01}\\[0.1cm]
        \end{tabular}
  }
  \hfill
  \subfloat[GSS\_IFV\bf{(84.63)}\label{tab:chapter4:1f}]{
    \small
    \scriptsize
    \begin{tabular}{@{\ }l@{\ } | @{\ }l@{\ } | @{\ }l@{\ } | @{\ }l@{\ } | @{\ }l@{\ } | @{\ }l@{\ } | @{\ }l@{\ }}
        \% & Cen. & Gol. & Hom. & Nuc. & NuM. & Spe. \\
        \hline \hline
        Cen. & \bf{96.95} & 0.14 & 0.16 & 1.13 & 0.17 & 1.44\\[0.1cm]
        Gol. & 0.20 & \bf{70.73} & 9.08 & 9.08 & 9.83 & 1.09\\[0.1cm]
        Hom. & 0.23 & 0.46 & \bf{78.94} & 4.67 & 7.18 & 8.52\\[0.1cm]
        Nuc. & 0.87 & 0.47 & 1.55 & \bf{94.05} & 1.74 & 1.32\\[0.1cm]
        NuM. & 0.09 & 0.30 & 5.28 & 0.69 & \bf{91.49} & 2.13\\[0.1cm]
        Spe. & 7.74 & 0.44 & 12.70 & 1.94 & 1.57 & \bf{75.60}\\[0.1cm]
        \end{tabular}
  }
\label{tab:confusingmatrics}
\end{table*}


\subsection{Experimental Results on the Test Data Set}
All results reported in this subsection are provided by the organizer of previous contests: the ICIP 2013 and ICPR 2014 contests. In this part, we compared our GSS\_IFV with six well-performing methods. These approaches are listed as follows:

\begin{itemize}
\item[--] Shen et al.~\cite{icip2013contest} combined the PRICoLBP~\cite{qi2014pairwise} and the Bag of SIFT~\cite{csurka2004visual}, and used a linear SVM classifier.

\item[--] Vestergaard et al.~\cite{larsen2014hep} proposed a type of shape index histograms (SIH) with donut-shaped spatial pooling for the cell classification task. The computational complexity of the SIH method is extremely low.


\item[--] Paisitkriangkrai et al.~\cite{icpr2014contest} applied a multi-class Boosting~\cite{paisitkriangkrai2014scalable} approach to automatically recognize different patterns of the HEp-2 cells. In their system, they used five types of features.
\item[--] Gao et al.~\cite{gao2014hep} utilized the Convolutional Neural Network (CNN)~\cite{lecun1998gradient}, and used a seven-layers CNN that consisted of three convolution layers, three pooling layers and one fully connection layer.
\item[--] Theodorakopoulos et al.~\cite{theodorakopoulos2014hep} combined a set of morphological features which contained two dimensional Boolean texture models and several textural descriptors.
\item[--] Sansone et al.~\cite{gragnaniello2014biologically} used a rotation invariant dense local descriptor ~\cite{kokkinos2012dense} with kernel codebook soft assignment under the BoW model.
\end{itemize}




Table~\ref{tab:confusingmatrics} shows the classification confusion matrices of our approach and six other methods.
Note that the winner (87.1\%)~\cite{manivannan2014hep} of ICPR 2014 contest used 5,000 additional images from the Task-2 along with all 13,596 training images. We do not list their result in Table~\ref{tab:confusingmatrics} because it is unfair to compare all relevant methods that only used the provided training data with the approach in~\cite{manivannan2014hep}.
Our method only used one type of feature, but many above-mentioned approaches combined multiple features.



We can observe that from the Table~\ref{tab:confusingmatrics}: 1). Among all the seven methods, our approach obtains the best averaged performance. Our approach improves Sansone's method by about 0.99 percentage point that is significant considering the difference (0.31 percentage point) between Sansone's (the second place) and Theodorakopoulos's (the third place). Meanwhile, we can also see that
  the proposed approach works best on four categories including ``Centromere'', ``Homogeneous'', ``Nucleolar'' and ``Speckled''. 2). The most confusing pairs are ``Homogeneous'' and ``Speckled'',  ``Golgi'' and ``Homogeneous'', and ``Golgi'' and ``Nuclear Membrane''. The reason why ``Homogeneous'' and ``Speckled'' are easily misclassified into each other is that these two categories have similar shape.


\section{Conclusion and Discussion}
\label{sec:conclusion}
In this paper,
we study the role of Gaussian Scale Space (GSS) theory as a pre-processing approach for HEp-2 cell classification task, and evaluate the GSS under two frameworks: the LBP and the BoW frameworks. Before, most research works on HEp-2 cell classification focused on feature extraction, feature encoding and classifiers. Very few efforts have been devoted to study the importance of the pre-processing. The proposed approach, using only one type of feature (LOAD), achieves superior performance on the large scale test data set maintained by the organizers. The proposed pre-processing approach can be generalized to most of the existing works~\cite{nosaka2014hep, qi2014pairwise, manivannan2014hep}, especially the BoW-based and LBP-based approaches. We also expect that the proposed GSS pre-precossing approach can be applied to the deep learning approach as a data augmentation technique.



\label{sec7}

\section*{Acknowledgements}
The authors would like to thank the organizers of the HEp-2 cell classification contests. This work was supported by the Academy of Finland and Infotech Oulu.





\bibliographystyle{elsarticle-num}

\bibliography{sample}

\end{document}